\documentclass[twocolumn]{article}

\usepackage{PRIMEarxiv}

\usepackage[utf8]{inputenc} 
\usepackage[T1]{fontenc}    
\usepackage{hyperref}       
\usepackage{url}            
\usepackage{booktabs}       
\usepackage{amsfonts}       
\usepackage{nicefrac}       
\usepackage{microtype}      
\usepackage{lipsum}
\usepackage{fancyhdr}       
\usepackage{graphicx}       
\usepackage{algorithm}
\usepackage{algorithmic}
\usepackage{multicol}
\usepackage{tabularx}
\usepackage{xcolor}
\usepackage{color}
\usepackage{textcomp}
\usepackage{multirow}
\usepackage{flushend}
\usepackage{abstract}

\usepackage[caption=false,font=footnotesize,labelfont=sf,textfont=sf]{subfig}

\usepackage{ulem}
\usepackage[nottoc]{tocbibind}
\usepackage{etoolbox}

\usepackage{amssymb, amsfonts, amsmath, amsthm}
\usepackage[justification=centering]{caption}

\newtheorem{theorem}{Theorem}

\graphicspath{{media/}}     
\title{LAC: Graph Contrastive Learning with Learnable Augmentation in Continuous Space
\thanks{\textit{\underline{Citation}}: 
\textbf{Authors. Title. Pages.... DOI:000000/11111.}} 
}

\author{
Zhenyu~Lin,
        Hongzheng~Li,
        Yingxia~Shao,
        Guanhua~Ye,
        Yawen~Li,
        and~Quanqing~Xu \\
School of Computer Science\\
Beijing University of Posts and Telecommunications\\
Haidian, Beijing, 100088.\\
\texttt{\{linzy, Ethan\underline{ }Lee, shaoyx, g.ye\}@bupt.edu.cn, warmly0716@126.com, xuquanqing.xqq@oceanbase.com}\\
}

\begin{document}
\twocolumn[
\begin{@twocolumnfalse}
\maketitle 
\begin{abstract}
    Graph Contrastive Learning frameworks have demonstrated success in generating high-quality node representations. 
    The existing research on efficient data augmentation methods and ideal pretext tasks for graph contrastive learning remains limited, resulting in suboptimal node representation in the unsupervised setting.
    In this paper, we introduce LAC, a graph contrastive learning framework with learnable data augmentation in an orthogonal continuous space. To capture the representative information in the graph data during augmentation, we introduce a continuous view augmenter, that applies both a masked topology augmentation module and a cross-channel feature augmentation module to adaptively augment the topological information and the feature information within an orthogonal continuous space, respectively. The orthogonal nature of continuous space ensures that the augmentation process avoids dimension collapse. 
    To enhance the effectiveness of pretext tasks, we propose an information-theoretic principle named InfoBal and introduce corresponding pretext tasks. These tasks enable the continuous view augmenter to maintain consistency in the representative information across views while maximizing diversity between views, and allow the encoder to fully utilize the representative information in the unsupervised setting. Our experimental results show that LAC significantly outperforms the state-of-the-art frameworks.
\end{abstract}
\end{@twocolumnfalse}
]

\section{Introduction}
\label{sec:introduction}

{Graph} Contrastive Learning (GCL) \cite{b1,b2} enhances generalization performance by leveraging multiple augmented views to learn latent information in graph data.
GCL serves as a powerful tool to address challenges associated with label sparsity and unlabeled data. It alleviates the expensive data labeling cost while significantly enhancing the generalization capabilities of graph neural network models across diverse downstream tasks.

Typically, GCL frameworks typically consist of {pretext tasks}, view augmenters, and encoders. {The pretext tasks guide the view augmenter and encoder to utilize representative information related to downstream tasks. The view augmenters transform input graphs into multiple correlated augmented views \cite{AutoGCL}. The encoders get representations from different augmented views.} 
The augmenters and encoders mutually influence each other \cite{GPA}. Specifically, high-quality augmented views contribute to the training of more powerful and generalized encoders \cite{MVGRL, JOAO, MERIT}, and the quality of encoders assists in learning effective augmenters as well.

GCL plays a pivotal role in various domains, including {emotion recognition \cite{AMGCT, MV-SSTMA}, multi-modal recommendation systems \cite{LATTICE, GRCN, FREEDOM}, and anomaly detection \cite{b4,b5,b6}.} Despite these capabilities, existing GCL frameworks still encounter two significant challenges in the unsupervised setting:

\textbf{Existing augmentation methods are not sufficient.} View augmenters in GCL frameworks fall into two categories: manual augmentation and learnable augmentation. The manual augmentation \cite{MVGRL, MERIT, GraphCL, BGRL, GCA, GRACE, SUGRL, GCC} involves selecting augmenters and their hyperparameters empirically from predefined options through numerous trial-and-error experiments per dataset \cite{JOAO}. The learnable augmentation \cite{GPA, ADGCL, AutoGCL, GraphLP, NCLA, GraphCLA, AdaMIP, SpCo, SPAN} automatically learns from data and generates augmented views. Although this approach avoids the laborious \cite{GraphLP} and time-consuming \cite{GraphCLA} process of numerous trial-and-error experiments, it augments the original graph data discretely, yielding {non-ideal} augmented views.
This {non-ideal} situation lies in both topology augmentation and feature augmentation. 
Firstly, the discrete perturbation of topological information leads to {non-ideal} views. 
Even the slightest perturbation to the topology such as removing an edge or dropping a node can destroy the representative information \cite{AutoGCL, GraphCLA} in the graph. 
For example, in chemical molecular or atomic classification tasks, where key chemical bonds are representative information \cite{SPAN}, random removal of these edges produces a low-quality view. 
Secondly, the discrete perturbation of continuous feature information also leads to {non-ideal} views. 
Initially, node features are represented in a continuous high-dimensional space, with each dimension containing rich information. However, existing works adopt a discrete approach to feature augmentation. For example, some work \cite{JOAO, AutoGCL, AdaMIP} randomly mask the specific dimension of all node features. The discrete augmentation for features also destroys the representative information in the graph data and may cause dimensional collapse \cite{SFA, MC-DCD}.

\textbf{Pretext tasks are not effective.} Existing studies \cite{MC-DCD, AutoGCL, ADGCL} emphasize that an effective pretext task should maintain consistency in the representative information across multiple augmented views, while ensuring the diversity of these views to prevent the model from generating overly similar views and embeddings \cite{MC-DCD}.
Suppose there is a graph-based multi-modal recommendation task \cite{LATTICE, GRCN, FREEDOM}. The multimodal features of items are extracted from their contents, which include visual and textual information. 
For example, the visual representative information that effectively distinguishes items may be about the shape and color of the item, rather than other attributes such as the visual background. The textual representative information of items may be brand and title instead of size information. 
When applying GCL algorithms, an effective pretext task must keep the different augmented views consistent with representative information about the item's shape, color, brand and title, while maximizing diversity by including as many different backgrounds and sizes as possible to improve the generalization ability of the model. 
If the textual and visual features contain the same semantic information (i.e. text description about the item's visual color), the consistency of cross-modal representative information should be maintained as well.
However, several studies based on the InfoMin principle \cite{ADGCL, AutoGCL, GraphLP, GraphCLA} neglect consistency constraints, leading to a loss of representative information during graph data augmentation \cite{ADGCL}. Other research \cite{NCLA} ignores diversity constraints, producing overly similar augmented views that result in shortcut solutions. 
This ultimately impairs the performance of GCL frameworks \cite{ADGCL, GIB}.

To overcome the above two challenges, we propose LAC, a novel Graph Contrastive Learning framework featuring \textbf{L}earnable \textbf{A}ugmentation in \textbf{C}ontinuous space. 
To address the first challenge, 
we propose a learnable Continuous View augmenter (CVA) in LAC, which identifies an orthogonal continuous Space by spectral theorem \cite{LinearAlgebra} and augments topological and feature information in this space.
We propose a Masked Topology Augmentation (MTA) module and a Cross-channel Feature Augmentation (CFA) module in CVA to search for appropriate augmented information.
To address the second challenge, we introduce a principle named Information Balance (InfoBal) for both the augmenter and the encoder in LAC. InfoBal's two sub-principles ensure consistency in the representative information among augmented views and while maximizing diversity. Furthermore, we introduce two pretext tasks derived from the two sub-principles for the training of augmenter and encoder.
Experiments are carried out on seven public datasets to validate the effectiveness of LAC compared to state-of-the-art (SoTA) methods. 
In summary, our contributions are as follows:
\begin{itemize}
\item We propose a graph contrastive learning framework called LAC, which includes a learnable continuous augmentation method and effective pretext tasks.

\item We design CVA, a learnable augmentation module that augments graph data in orthogonal continuous space to generate more ideal augmented views and avoid dimensional collapse.  

\item We introduce the InfoBal principle and design two pretext tasks based on it to guide view augmenters in maximizing diversity and ensuring consistency of representative information across views, while helping encoders fully utilize representative information.

\item  Experimental results on seven sparse datasets show that LAC outperforms SoTA GCL frameworks in an unsupervised setting. 
\end{itemize}

\section{Preliminaries}
\label{sec:Preliminaries}

\subsection{Notions} 
An undirected and connected graph is denoted as $\mathcal{G}=(\mathcal{V},\mathcal{E})$, where $\mathcal{V}$ represents the node set and $\mathcal{E} \subset \mathcal{V} \times \mathcal{V}$ represents the edge set. The topological adjacency matrix is $\mathbf{A}\in \mathbb{R}^{N\times N}$, the node feature matrix is $\mathbf{X} \in \mathbb{R}^{N\times d}$, and $\mathbf{x}_i$ denotes the feature of node $i$, $\forall i\in \mathcal{V}$. The normalized Laplacian matrix of the graph $\mathcal{G}$ is given as $\tilde{\mathbf{L}}=\mathbf{I}-\tilde{\mathbf{D}}^{-1/2}\tilde{\mathbf{A}}\tilde{\mathbf{D}}^{-1/2}$, where $\tilde{\mathbf{D}}$ denotes the degree matrix of the normalized adjacency matrix $\tilde{\mathbf{A}}$. In contrastive learning, each graph can also be denoted as view V.

{\subsection{Graph Contrastive Learning} The objective of graph contrastive learning is to distinguish between latent representative information in the data \cite{SimCLR} and to learn embeddings for downstream tasks. 
View augmenters can either be manually selected from predefined pools \cite{MVGRL, MERIT, GCA, BGRL, GRACE, SUGRL, GCC} or designed as learnable and generative \cite{GPA, JOAO, ADGCL, AutoGCL, GraphLP, NCLA, GraphCLA, AdaMIP}. 
The augmenters generate one or multiple augmented views. 
For example, given original data, data augmenters $g_1, g_2$, original data view $V$, the augmented view $V'$ can be denoted as:
\begin{equation}
    V' = (A', X') = g(V) = g((A,X)).
\end{equation}
Subsequently, based on the positive and negative sample pairs defined across multiple views, GCL frameworks train the encoders to obtain embeddings for downstream tasks. We denote encoders as $f_1, f_2$, InfoNCE loss~\cite{InfoNCE} as I. {The unsupervised GCL framework often uses pretext tasks based on the InfoMax~\cite{InfoMax} and InfoMin~\cite{InfoMin} principle to train encoders and augmenters respectively, which can be expressed as:}
\begin{equation}
\begin{split}
\mathop{\min} I(f_1(g_1(V)), f_2(g_2(V))), fix \ f_1, f_2, \  \\
s.t. \mathop{\max} I(f_1(g_1(V)), f_2(g_2(V))), fix \ g_1, g_2.
\end{split}
\end{equation}

\subsection{Theorems}
In linear algebra, the spectral theorem \cite{LinearAlgebra} establishes a standard framework for decomposition vector space and gives the condition that a matrix can be diagonalized.
\begin{theorem} \label{Theorem:1}
\textbf{(Spectral Theorem).} Let P be a symmetric matrix on $\mathbb{R}^{n \times n}$, then P can be decomposed into:
\begin{equation}
    \mathbf{P} = \mathbf{U} \mathbf{\Lambda} \mathbf{U}^T, 
\end{equation}
\textit{where $\mathbf{U} \in \mathbb{R}^{n \times n}$ is a set of orthogonal eigenvectors of  $\mathbf{P}$, $\mathbf{\Lambda} \in \mathbb{R}^{n \times n}$ is the corresponding diagonal matrix of eigenvalues, and $\mathbf{U}^T$  is the transposed matrix of $\mathbf{U}$, satisfying:}
\begin{equation}
    \mathbf{U} \mathbf{U}^T = \mathbf{I}_n,
\end{equation}
\textit{where $\mathbf{I}_n \ \in \mathbb{R}^{n \times n}$ is an identity matrix.} 
\end{theorem}

\begin{figure*}[!htp]
    \centering
    \includegraphics[width=1\linewidth]{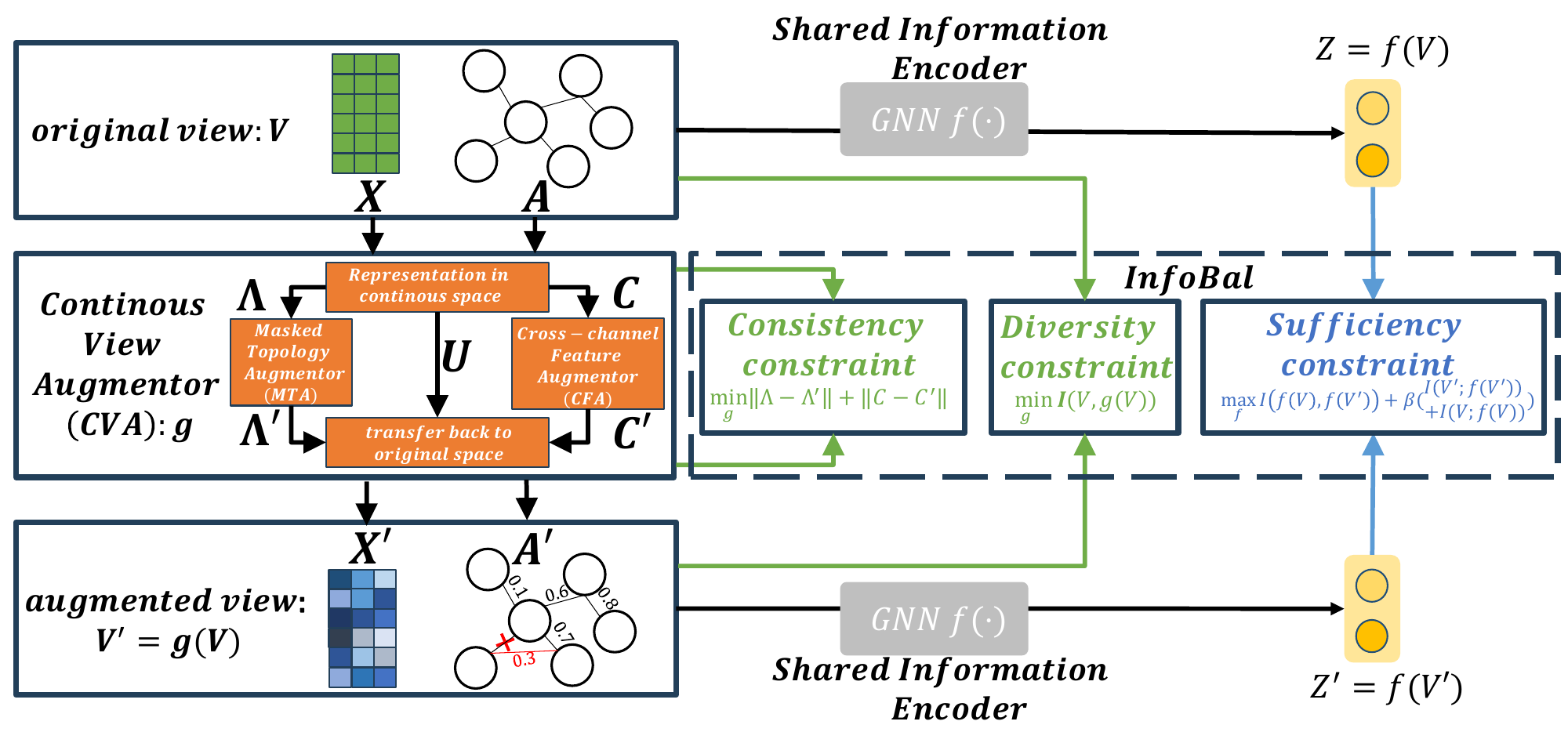}
    \caption{The overview of LAC.}
    \label{fig:1}
\end{figure*}

In approximation theory~\cite{LinearAlgebra}, an arbitrary complicated multivariate function can be approximated by a series of simple univariate functions.
\begin{theorem} \label{Theorem:2}
\textbf{(Kolmogorov–Arnold Theorem)~\cite{DeepSets}.} An arbitrary multivariate function $f : [0, 1]^N \rightarrow \mathbb{R} (N > 0)$ can be modeled using the following expression :
\begin{equation}
    f(x_1, ... , x_N) = \rho (\sum_{p=1}^N \alpha_{i,p} \phi(x_p)),
\end{equation} where $\rho : \mathbb{R}^d \rightarrow \mathbb{R}$, $\phi : \mathbb{R} \rightarrow \mathbb{R}^d, d>0 $ and $\alpha_{i,p}$ is the weight.
\end{theorem}

\begin{theorem} \label{theorem:3}
\textbf{(Information Inequality).}
\textit{The upper bound of mutual information between the representations of the augmented view $V'$ and original view $V$ is given by:}
\begin{equation}
I(f(V);f(V')) \leq min \{I(V; V'), I(V';f(V')), I(V; f(V))\}, 
\end{equation}
where $f$ is an information encoder.
\end{theorem}

\begin{proof} We have a Markov chain: $V \rightarrow V'=g(V) \rightarrow f(g(V))$, the first arrow means augmentation, and the second arrow denotes information encoding. According to the data process inequality \cite{ElementsOfInformationTheory}, we infer that:
\begin{equation}
I(V; f(V')) \leq min \{I(V; V'), I(V';f(V')\}. 
\end{equation}
Meanwhile, there is another Markov chain: $f(V) \leftarrow V \rightarrow f(V')$, which is Markov equivalent to $f(V) \rightarrow V \rightarrow f(V')$ since $f(V)$ and $f(V')$ are conditionally independent after observing $V$. Thus, the following equation holds:
\begin{equation}
\begin{split}
I(f(V);f(V')) &\leq min \{I(V; f(V')), I(V; f(V)) \} \\
          &\leq min \{I(V; V'), I(V';f(V')), I(V; f(V))\}. 
\end{split}
\end{equation}
Above all, Theorem~\ref{theorem:3} is proven.
\end{proof}


\section{LAC Framework Overview}

Figure~\ref{fig:1} provides an overview of LAC, which includes the Continuous View Augmenter, the Shared Information Encoder, and the pretext tasks built on the InfoBal. 

\textbf{Continuous View Augmenter (CVA)}. The CVA augments both topological and feature information in an orthogonal continuous space, generating augmented views accordingly. In CVA, we propose the Masked Topology Augmentation module for topology augmentation and the Cross-channel Feature Augmentation module for feature augmentation, respectively.
The continuous augmentation method employed by the CVA minimizes information loss typically associated with discrete augmentation methods, thereby enhancing suitability for graph data. The detailed design of CVA is presented in Section~\ref{sec:cva}.

\textbf{Shared Information Encoder}. The shared information encoder extracts information from the original view and the augmented view to learn the node embeddings. The encoder $f$ comprises a K-layer Graph Neural Network (GNN). The shared information encoder is a common and widely adopted technique in existing works~\cite{GRACE, GCA}. The embeddings $\mathbf{Z}$ for nodes in view $V$ and the embeddings of nodes in augmented views $\mathbf{Z}^{'}$ are yielded by shared information encoder $f$:
\begin{equation}
\mathbf{Z} = f(V) = f(\mathbf{A}, \mathbf{X}), 
\mathbf{Z}^{'} = f(V') = f(\mathbf{A}^{'}, \mathbf{X}^{'}).
\end{equation}

\textbf{InfoBal Pretext Task}.
To achieve high-quality node representations in unsupervised scenarios, the InfoBal pretext task is proposed. The InfoBal framework comprises two components.
Firstly, it introduces diversity, and consistency constraints for the CVA, and guides the CVA to generate appropriate augmented views with adequate augmentation variances while preserving representative information.
Secondly, InfoBal imposes a sufficiency constraint on the training of the shared encoder.
This helps the shared encoder to extract more representative information from views.
In contrast to InfoMin~\cite{InfoMin}, which is typically employed for training augmenters, InfoBal emphasizes the balance of consistency and diversity among augmented views.
As opposed to InfoMax~\cite{InfoMax}, traditionally used for encoder training, InfoBal boosts the sufficiency of utilizing representative information in augmented views.
Thus, InfoBal allows the model to generate high-quality node representations in the unsupervised scenario. The detailed design of InfoBal is presented in Section~\ref{sec:cva}.

In the InfoBal pretext tasks, we use InfoNCE \cite{InfoNCE} to measure the mutual information between two views. 
We regard the embeddings of same node in the original view $V$ and augmented view $V'$ as positive pairs (e.g., ($\mathbf{z}_{i}, \mathbf{z}_{i}^{'}$ $\vert$ $i \in \mathcal{V}$)). On the other hand, we consider the embeddings of different nodes in the two views as negative pairs (e.g., ($\mathbf{z}_{i}, \mathbf{z}_{k}^{'}$ $\vert$ $i,k \in \mathcal{V}, i \neq k$)). We define the mutual information extracted from two views by encoder as follows:
\begin{equation}
     I(f(V); f(V')) =\frac{1}{2N} \sum_{i=1}^{N}log{\frac{{e^{s(\mathbf{z}_i,\mathbf{z}_{i}^{'})/ \tau}}}{{\sum_{k\in \mathcal{V}}{{e^{s(\mathbf{z}_i,\mathbf{z}_{k}^{'})/ \tau}}}}}}.
\end{equation}
The similarity between two embeddings of nodes is measured by the cosine function $s(\cdot)$. The temperature parameter $\tau$ controls the sharpness of the similarity scores. 

Finally, the augmenter and encoder are trained against each other, and the loss function is expressed as follows:

\begin{equation} 
\begin{split}
    \mathop{\min}\limits_{g}{{I}_{diversity}} + \alpha {I}_{consistency}. \\
     s.t. \mathop{\max}\limits_{f}{{I}_{sufficiency}} .
\end{split}
\end{equation}

\section{Continuous View Augmenter}
\label{sec:cva}
In this section, we introduce the Continous View Augmenter (CVA). It first transforms the original graph data into the representations in an orthogonal continuous space represented by the bases $\mathbf{U}$. On basis of the continous representation, CVA applies Masked Topology Augmentation (MTA) and Cross-channel Feature Augmentation (CFA) to generate augmented topology and feature of the graph data, respectively. Finally, CVA transforms the augmented results back to the original discrete space to get the final augmented views. 

\subsection{Representation in Orthogonal Continuous Space}
To represent information in a continuous space, establishing a coordinate system with orthogonal vector bases is critical. This approach allows topological data $\mathbf{A}$ and feature data $\mathbf{X}$ to be expressed as linear combinations of these bases, preserving data integrity and preventing dimensional collapse during perturbations.

\textbf{Topology representation}. The undirected graph topological information is denoted as a symmetric and real matrix $\mathbf{A}$. 
According to the Theorem~\ref{Theorem:1}, 
$\mathbf{A}$ can be decomposed into a set of orthogonal vector bases $\mathbf{U} \in \mathbb{R}^{N\times N}$ and a diagonal eigenvalue matrix $\mathbf{\Lambda}$ as follows: 
\begin{equation} \label{eq:lambda}
\mathbf{A} = \mathbf{U} \mathbf{\Lambda} \mathbf{U^T}, \mathbf{U^T}\mathbf{U}=I_N, 
\end{equation}
where the vector bases $\mathbf{U}$ are the \textbf{\textit{orthogonal continuous space}}. The representation of topological information $\mathbf{A}$ in the continuous space associated with $\mathbf{U}$ is $\mathbf{U}^T \mathbf{A}$ according to the definition of linear mapping~\cite{LinearAlgebra}. 

Augmenting topological information in a continuous space presents a significant challenge due to the high computational complexity. The standard approach results in a complexity of $\mathcal{O}(n^2)$. To address this issue, we utilize a key characteristic of the matrix product $\mathbf{U}^T \mathbf{A}$ to significantly reduce the computational overhead during the augmentation process. We express it as follows:
\begin{equation}
    \mathbf{U}^T \mathbf{A} = \mathbf{U}^T \mathbf{U} \mathbf{\Lambda} \mathbf{U^T} = \mathbf{\Lambda} \mathbf{U^T}.
\end{equation}
Considering that the basis of continuous space $\mathbf{U}$ is fixed, this insight enables us to perturb $\mathbf{\Lambda}$ rather than $\mathbf{U}^T \mathbf{A}$. This alteration simplifies the augmentation process, and reduces the computational complexity from $\mathcal{O}(n^2)$ to $\mathcal{O}(n)$, marking a substantial improvement.

\textbf{Feature representation}. The representation of $\mathbf{X}$ in the orthogonal continuous space is denoted as $\mathbf{U}^T \mathbf{X}$. 
However, $\mathbf{U}^T \mathbf{X}$ does not have a property similar to $\mathbf{U}^T \mathbf{A}$, and we cannot simplify the perturbation of $\mathbf{U}^T \mathbf{A}$. For convenience, we use $\mathbf{C}$ to denote the representation of feature information in continuous space as follows:  
\begin{equation}
    \mathbf{C} = \mathbf{U}^T \mathbf{X}.
\end{equation}

\subsection{Masked Topology Augmentation}
MTA augments the topological information represented by the $\mathbf{\Lambda}$ matrix in continuous space.
It's worth noting that the topological information $\mathbf{\Lambda}$ exhibits two unique properties among the eigenvalues on its diagonal:
(1) The eigenvalues $\lambda_i, \forall i \in [1,N]$ along the diagonal of the $\mathbf{\Lambda}$ matrix are within the range of [-1,1], each containing crucial topological information; (2) The difference between adjacent eigenvalues  $\Delta \lambda_i = \lambda_{i+1} - \lambda_i$ encapsulates key topological insights \cite{Specformer}.

On the basis of the above two properties, it is crucial to perturb both eigenvalues and their interdependencies collectively when augmenting topological information. The relationship between eigenvalues satisfies a multivariate function. According to the Theorem~\ref{Theorem:2}, each augmented $\lambda_{i}^{'} (i \in [1,N])$ can be denoted by a multivariate function $\psi_i$:
\begin{equation}
    \lambda_{i}^{'} = \psi_i(\lambda_1, ... , \lambda_N) = \rho (\sum_{p=1}^N \alpha_{i,p} \phi(\lambda_p)),
\end{equation}
{where $\rho : \mathbb{R}^d \rightarrow \mathbb{R}$, $\phi : \mathbb{R} \rightarrow \mathbb{R}^d$, $\alpha_{i,p}$ represents the weight of $\phi(\lambda_p)$ for $ d \textgreater 0$.}

In this paper, we introduce a transformer-based MTA module to implement these functions $\psi_i$. Figure~\ref{fig:2} illustrates that MTA comprises an eigen embedder, an encoder, and a decoder. As stated in Theorem~\ref{Theorem:4} from Section~\ref{sec:discussion}, MTA can be trained to approximate any multivariate function.

\begin{figure}[!tbp]
    \centering
    \scalebox{0.7}{
        \includegraphics[width=1\linewidth]{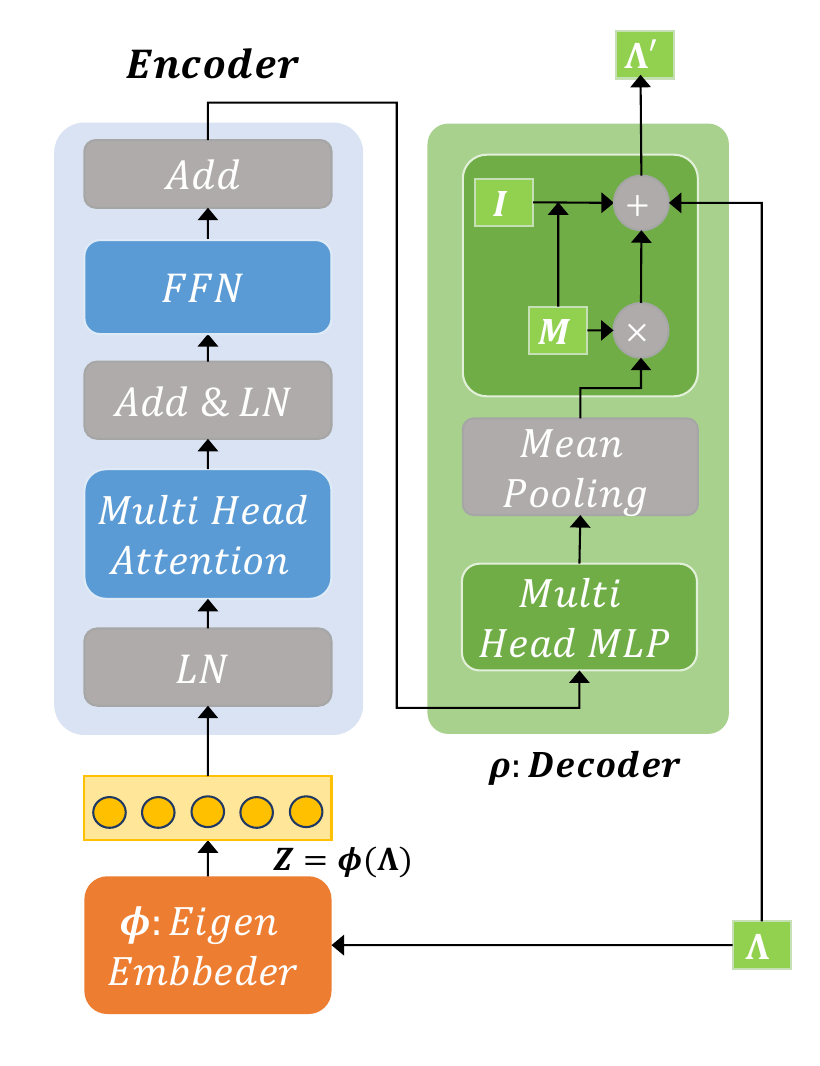}
    }
    \caption{\centering{The structure of MTA.}}
    \label{fig:2}
\end{figure}

\textbf{Eigen Embedder}. The eigen embedder $\phi$ maps each $\lambda_i$ to a $d$-dimensional embedding $\phi(\lambda_i)$. The $j$-th value of the embedding $\phi(\lambda_i)$ is:
\begin{equation}\label{eq:11}
    \phi(\lambda_i)_j =  \left\{
\begin{aligned}
& \lambda_i, \quad j=1, \\
& sin(\frac{\epsilon \lambda_i}{N^{2k/d}}), \quad  j=2k \\
& cos(\frac{\epsilon \lambda_i}{N^{2k/d}}), \quad j=2k+1 \\
\end{aligned}
\right. ,
\end{equation}
where $\epsilon$ is a hyperparameter, and \( k \) is the floor function applied to \( j/2 \). This type of embedding represents the differences between eigenvalues in a way similar to the positional embedding \cite{Specformer} which is usually used in transformer architectures. Consequently, we get the initial high dimensional embeddings of $\mathbf{\Lambda}$: 
\begin{equation}
    \mathbf{Z} = [\phi(\lambda_1), ..., \phi(\lambda_N)]^T \in \mathbb{R}^{N \times d}.
\end{equation}

\textbf{Encoder}. The encoder is designed for capturing the underlying relationship among the $\phi(\lambda_i), i \in [1, N]$. Each encoder layer consists of two main components: a multi-head self-attention (MHA) module with residual connections and a feedforward network (FFN) with residual connections. Layer normalization (LN) is applied before feeding the representation into each module \cite{preLN}. Formally, each embedded vector $\phi(\lambda_i)$ treated as a concatenation of $H$ vectors across different heads:
\begin{equation}
    \mathbf{Z}_i = \phi(\lambda_i) = \Vert (\phi_1(\lambda_i), ..., \phi_H(\lambda_i)), i \in [1,H],
\end{equation}
{where $\Vert$ means concatenation operation, $\phi_i: \mathbb{R} \rightarrow \mathbb{R}^{d/H}, i \in [1,H]$. In other words, we divide whole $\mathbf{Z}$ into $H$ heads. The calculation of each head is as follows:}
\begin{equation} 
    \tilde{\mathbf{Z}}_h^{l} = MHA(LN(\hat{\mathbf{Z}}_h^{l-1})) + \hat{\mathbf{Z}}_h^{l-1}, \hat{\mathbf{Z}}_h^{0} = \mathbf{Z}_h ,
\end{equation}
\begin{equation}
    \hat{\mathbf{Z}}_h^{l} = FFN(LN(\tilde{\mathbf{Z}}_h^{l})) + \tilde{\mathbf{Z}}_h^{l} .
\end{equation}
\begin{equation} \label{eq:17}
    \hat{\mathbf{Z}}_{h}^{L} = [\sum_{p=1}^{N}\alpha_{1,p}^{h}\phi_h(\lambda_p), ..., \sum_{p=1}^{N}\alpha_{N,p}^{h}\phi_h(\lambda_N)]^T, 
\end{equation}
{where $\alpha_{i,p}^{h}$ means the weight of $\phi_h(\lambda_i)$ and $\phi_h(\lambda_p)$. The Eq. \eqref{eq:17} illustrates that the function of the encoder of MTA is to calculate weighted aggregation of $\mathbf{Z}$.}

\textbf{Masked Multi-head Decoder}. The decoder {$\rho$} decodes the high dimensional topological information into the augmented diagonal matrix $\mathbf{\Lambda^{'}}$. The output $\hat{\mathbf{Z}}_h^{L}$ from the encoder is then decoded by the multi-head MLP, resulting in the $h$-th perturbed eigenvalues matrix $\mathbf{\Lambda}^{'}_h$:
\begin{equation}
    \mathbf{\Lambda}_{h}^{'} = diag(\sigma( \hat{\mathbf{Z}}_h^{L} \mathbf{W}^h)),
\end{equation}
where $\sigma$ represents the activation function and $\mathbf{W}^h \in \mathbb{R}^{d/H \times 1}$. The augmented matrix $\mathbf{\Lambda^{'}}$ is derived by performing mean pooling across the eigenvalues matrices obtained from attention heads:
\begin{equation}
    \mathbf{\Lambda}^{'} = Mean({\mathbf{\Lambda}_{h}^{'} | \forall h \in [1, H]}).
\end{equation}

To enhance the quality of augmentation outcomes, we introduce an adjustable diagonal mask matrix $\mathbf{M} =diag(m_1, ..., m_N) \in \mathbb{R}^{N \times N}$ used to mask a specific range of eigenvalues on the $\mathbf{\Lambda^{'}}$ matrix. The ultimate $\mathbf{\Lambda}^{'}$ is denoted as follows:
{
\begin{equation} 
\begin{split}
    \mathbf{\Lambda}^{'} 
                    &= (\mathbf{I}-\mathbf{M}) \odot \mathbf{\Lambda} + \mathbf{M} \odot \mathbf{\Lambda}^{'} \\
                    &= diag((1-m_1)\lambda_1+m_1*\sigma(\frac{1}{H}\sum_{h=1}^{H} \sum_{p=1}^{N} \alpha_{1,p}^{h}\phi_{h}(\lambda_{1})W_h), \\
                    & ..., (1-m_N)\lambda_N+m_N*\sigma(\frac{1}{H}\sum_{h=1}^{H} \sum_{p=1}^{N} \alpha_{N,p}^{h}\phi_{h}(\lambda_{N})W_h)), \\
                    &= diag(\rho (\sum_{p=1}^N \alpha_{1,p} \phi(\lambda_p)), ... , \rho (\sum_{p=1}^N \alpha_{N,p} \phi(\lambda_p))). 
\end{split}
\end{equation}}
Here, $\mathbf{I} \in \mathbb{R}^{N \times N}$ denotes the identity matrix, and $\odot$ indicates element-wise multiplication between matrices.

\begin{figure}[!t]
    \centering
    \scalebox{0.8}{
        \includegraphics[width=1\linewidth]{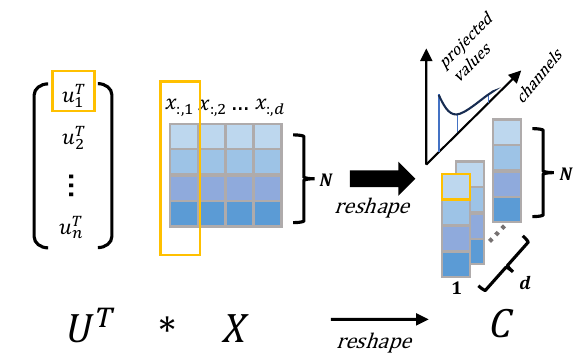}
    }
    \caption{The projection of node features.}
    \label{fig:3}
\end{figure}

\subsection{Cross Channel Feature Augmentation} 
The continuous representation of feature information $\mathbf{C}=\mathbf{U}^T\mathbf{X}$ is modeled as the coordinate coefficient of $\mathbf{X}$ in the orthogonal continuous space represented by $\mathbf{U}$, as depicted in Figure~\ref{fig:3}. Specifically, $c_{i,j}$ represents the projected value of the $j$-th channel graph signal $\mathbf{x}_{:,j}$ onto basis $\mathbf{u}_i$. Each column of $\mathbf{C}$ delineates the distribution of projected values for a channel of the graph signal across various bases. 

Randomly masking specific node features or feature dimensions can result in \textbf{\textit{dimensional collapse}} \cite{collapse, SFA}. To analyze this phenomenon, we provide empirical results on the Cora. As shown in Figure~\ref{fig:5}, the blue line represents the coefficients $C$, which exhibits a long-tail distribution. The imbalance of the distribution of projected values causes a few channels to dominate the feature information in the continuous space \cite{SFA}. For example, the projected values across certain channels in $\mathbf{C}$ are very small, while the projected values across other channels are very large. 
Consequently, these former channels become negligible in their capacity to describe information compared to the latter channels, leading to their invalidation. This causes a phenomenon that the high-dimensional representations of node features in continuous space collapse \cite{collapse} into low-dimensional representations, which makes it difficult to distinguish between different nodes \cite{SFA}.

To avoid the dimensional collapse problem, cross channel feature augmentation (CFA) is introduced. This method adaptively perturbs $\mathbf{C}$ using a cross-channel convolution $g_F$, which automatically captures the relationships {of projected values} across different channels and generates properly augmented features with balanced distributions.
We begin by reshaping $\mathbf{C} \in \mathbb{R}^{N \times d}$ into the $N \times d \times 1$ matrix as the input of CFA. Thus the shape of embeddings for each node is $d \times 1 $. 
We apply a convolution kernel $K \in \mathbb{R}^{d \times d \times 1}$ for each node embedding through the broadcast mechanism, producing an output of $d \times d \times 1$. This output represents a vector with $d$ channels, where each channel is a $d \times 1$ vector. Summing along the second dimension gives a new embedding of the node after merging different channel information, the shape is still $d \times 1$. A single convolution kernel $K$ is shared among all nodes. Finally, the augmented results of each node are concatenated to obtain $\mathbf{C}' \in \mathbb{R}^{N \times d} $, where the element $\mathbf{C}^{'}_{i,l}$ is denoted as follows:
\begin{equation}
    \mathbf{C}^{'}_{i,l} = g_{F}(\mathbf{C})_{i,l} = \sum_{j=1}^{D} C_{(i,j)} K_{(l,j,1)}.
\end{equation}

The existing random feature masking method is the special case of the proposed CFA (refer to Theorem~\ref{Theorem:4} in Section~\ref{sec:discussion} for details). 
In contrast to existing continuous feature perturbation methods \cite{COSTA, SFA}, CFA facilitates adaptive augmentation across diverse datasets, thereby reducing dependence on trial-and-error experimentation.
Ultimately, it enables CVA to adaptively generate more optimal node features for the augmented views.

\subsection{Discussion of CVA}  
\label{sec:discussion}

The $\mathbf{\Lambda}^{'}$ and $\mathbf{C}^{'}$ generated by the MTA and CFA cannot be directly applied in the GNN-based information encoder. We convert $\mathbf{\Lambda}^{'}$ and $\mathbf{C}^{'}$ from the continuous space back to the original space to obtain the final augmented view $V'$ as follows:
\begin{equation}
    V' = (\mathbf{A}^{'}, \mathbf{X}^{'}) =  (\mathbf{U}\mathbf{\Lambda}^{'}\mathbf{U}^T, \mathbf{U} \mathbf{C}^{'}) .
\end{equation}
Such augmented view can be utilized across various types of encoders.

\subsubsection{Relationship with the Augmentation in the Original Space}
We next analyze the connection between CVA and view augmentation in the original space.
In the original space, topological perturbations generate edges in $\mathbf{A}$ with integer weights of 0 or 1, while discrete feature perturbations mask random dimensions in $\mathbf{X}$ to zeros. Every discrete augmentation in the original space has a corresponding augmentation in our proposed continuous space, as proven by Theorem~\ref{Theorem:4}.

\begin{theorem} \label{Theorem:4}
    \textit{Any topological augmentation in discrete space and augmentation of discretized features has a corresponding augmentation on the topological and feature information in continuous space of LAC.}
\end{theorem}  

\begin{proof}  According to theoretical justification in \cite{SPAN}, the perturbation of $\mathbf{\Lambda}$ by flipping the edge between nodes p and q is given by the sum of all $\Delta \lambda_i = \lambda_i^{'} - \lambda_i$:
\begin{equation}
\sum_{i=1}^{N}|\Delta \lambda_i | = \sum_{i=1}^{N}|2 u_{i,p}u_{i,q} - \lambda_i (u_{i,p}^2 + u_{i,q}^2)|. 
\end{equation}
Consequently, perturbing any edge in the discrete space corresponds to a specific perturbation of the $\mathbf{\Lambda}$ matrix. 
Moreover, removing a node is equivalent to eliminating all edges associated with that node.

The overall perturbation on $\mathbf{\Lambda}$ is obtained by directly summing up all the perturbations of single edges on $\mathbf{\Lambda}$. For example, when removing node $s$, the perturbation to $\mathbf{\Lambda}$ is denoted as follows:
\begin{equation}
\sum_{t}^{\mathcal{N}_s}\sum_{i=1}^{N}|\Delta \lambda_i | = \sum_{t}^{\mathcal{N}_s}\sum_{i=1}^{N}|2 u_{i,s}u_{i,t} - \lambda_i (u_{i,s}^2 + u_{i,t}^2)|, 
\end{equation}
where $\mathcal{N}_s$ is the neighborhood of $s$. Thus, any perturbation of the topological structure in discrete space corresponds to a specific perturbation of  $\mathbf{\Lambda}$ in continuous space of LAC.

On the other hand, the random feature masking of node features $\mathbf{X}$ is equivalent to setting all the values in a dimension to zero. 
Specifically, if GCL frameworks mask the $l$-th dimension of features $\mathbf{X}$, it results in a matrix $\mathbf{X}_{:l}^{'}$ where the values in the $l$-th column are all set to 0. Since $\mathbf{X}_{:l}^{'} = (\mathbf{U}\mathbf{C}^{'})_{:l} = 0$, we conclude that the values in the $l$-th column of $\mathbf{C}^{'}$ are zero. CFA can achieve this by learning an appropriate set of convolutional kernel coefficients:
\begin{equation}
    \mathbf{C}^{'}_{i,l} = \sum_{j=1}^{D} C_{(i,j)} K_{(i,l,j)} = 0, \forall i \in [1,N].
\end{equation}

This demonstrates that any discretized feature perturbation is equivalent to perturbing C in the space introduced in LAC.

Therefore, any discrete augmentation, whether of topological structure or node features, has a corresponding representation in continuous space. Theorem~\ref{Theorem:4} is proven. 
\end{proof}

\subsubsection{Complexity Analysis}

\textbf{Time Complexity.} 
{The total time cost consists of preprocessing time and training time}. 
First, the preprocessing time is the time taken to decompose the $\mathbf{A}$ and $\mathbf{X}$ matrices. As CVA directly augments $\mathbf{\Lambda}$ and $\mathbf{C}$, the matrix decomposition is required only once during the preprocessing stage, rendering the preprocessing time effectively constant.
Second, the training time overhead includes two parts.
The first part comes from the computation of MTA and CFA in CVA. 
The encoder and decoder in MTA have $O(24ND^2+4DN^2)$ and $O(6NH)$ time complexity, respectively. The time complexity of a CFA is $O(Nd^2)$. $D$ is the hidden dimension, $d$ is the feature dimension, $H$ is the number of heads, and they are much small compared to the datasize $N$. Therefore, the total time complexity of the first part is $O(N^2)$. 
The second part of training time overhead comes from GNN encoding and contrastive learning, which are $O(N^2)$.

\textbf{Space Complexity.} The space overhead of LAC comes from two aspects. One is the storage overhead of $\mathbf{\Lambda}$, $\mathbf{U}$, and $\mathbf{C}$, which are $O(N^2)$. 
The other is the parameter storage overhead brought by the CVA and GNN modules, which is 
 independent of the dataset size. 
The parameters in MTA are $O(3LHD^2+2LD^2+HD)$. The parameters in CFA are $O(d^2)$. $L, H$ is the number of layers of the encoder, the number of heads in MTA, which is usually 1 or 2. $D$ is the hidden dimension of MTA, which is 128 or 256 in general. 
The space complexity of GNN is $O(dD+D^2)$.
Therefore, the total space complexity is $O(N^2)$. 

\section{The Pretext Tasks Based on InfoBal}
\label{sec:infobal}
In this section, we introduce a universal principle known as InfoBal. According to its sub-principles, we design two specific pretext tasks for the continuous view augmenter and the shared information encoder.

\subsection{InfoBal Principle}
The InfoBal principles enable the GCL framework to maintain consistency of representative information across multiple augmented views, while ensuring diversity, thereby facilitating efficient extraction of representative embeddings by the encoder.
Specifically, InfoBal adheres to two sub-principles that guide the augmenter and the encoder. The objectives of the two sub-principles are as follows: 1) the view augmenter should create diverse views (\textit{diversity constraint}) while maintaining representative information (\textit{consistency constraint}). 2) the encoder should extract as much representative information as possible from these views (\textit{sufficiency constraint}) to get embeddings while fulfilling the goal of maximizing mutual information.

\subsection{Pretext Task for CVA}
In semi-supervised contrastive learning, the InfoMin principle is usually used to train the augmenter. It optimizes the balance between view consistency and view diversity, extracting representative information from raw graph data with the help of labels \cite{InfoMin}. 
In unsupervised scenarios, extracting as much representative information as possible is equivalent to finding a set of views with maximal augmented variances (\textit{diversity constraint}) while preserving sufficient representative information (\textit{consistency constraint}). 
However, in the absence of labels, InfoMin, as utilized in unsupervised GCL frameworks \cite{ADGCL, AutoGCL, GraphLP, GraphCLA}, cannot guarantee the consistency of label-related information between augmented views when the encoders are trained using the InfoMax principle.
It causes the corruption of representative information in the augmented views during augmentation. Theorem \ref{theorem:5} formalizes the above problem:

\begin{theorem} \label{theorem:5}
     If the GCL framework adopts InfoMax principle $max \ I(f(V); f(V'))$, InfoMin principle $min \ I(V; V')$ to train the shared information encoder $f$ and the augmenter $g$ respectively, and the generated view will converge to a complete noise graph, making the representative information in the original graph data completely lost.
\end{theorem}

\begin{proof}  The optimal case of the $min \ I(V; V')$ is $I(V; V')=0$, i.e., the augmenter generates the augmented view without utilizing any information from the original data at all so that the augmented view does not contain any representative information. At the same time, according to Theorem~\ref{theorem:3}, the InfoMax principle for training encoder has:
\begin{equation}
\begin{split}
    I(f(V);f(V')) \leq   &I(V; f(V'))  \\ 
    &\leq  \min \{I(V; V'), I(V';f(V') \} \\
    & = min \{0, I(V';f(V') \} = 0,
\end{split}
\end{equation}
which illustrates that the level of utilization of the augmented information is limited by the level of utilization of the original information. Thus, when the optimal solution of the InfoMin principle is achieved—where the augmented view generated, $V'=g(V)$, satisfies $V' = \mathop{\min}\limits_{g}I(V;g(V))$—the augmenter is unable to utilize any information from the original data. Above all, Theorem \ref{theorem:5} is proven.
\end{proof}

Other works, such as those by \cite{GPA, JOAO, NCLA}, also face challenges, despite not employing the InfoMin principle. These studies overlook the need for diversity in augmentations, resulting in non-ideal views that resemble the original graph too closely and hindering the encoder's ability to discern representative information.

To address this issue, we integrate a consistency constraint with the diversity constraint as a pretext task in CVA to generate ideal augmented views. The consistency constraint uses MSE loss to measure the consistency between augmented values and original values. The diversity constraint is the minimization of mutual information between two views.

Assume view V is denoted as $(\mathbf{U}, \mathbf{\Lambda}, \mathbf{C})$ in LAC, the perturbation to view V can be regarded as the perturbation to $\mathbf{U}$, $\mathbf{\Lambda}$, and $\mathbf{C}$. Considering that $\mathbf{U}$ remains unchanged as the vector bases during the augmentation process, the diversity constraint compels LAC to maximally perturb $\mathbf{\Lambda}$ and $\mathbf{C}$.
Building on the diversity constraint, we introduce a regularized loss (\textit{consistency constraint}) to prevent excessive perturbation of $\mathbf{\Lambda}$ and $\mathbf{C}$. 
Therefore, the loss function of the pretext task is formulated as follows:

\begin{equation} 
\begin{split}
    \mathop{\min}\limits_{g}{\underbrace{I(V;g(V))}_{diversity}} + \alpha \underbrace{(\left\| \mathbf{\Lambda}- \mathbf{\Lambda}^{'} \right\|_F^2 + \left\| \mathbf{C} - \mathbf{C}^{'} \right\|_F^2)}_{consistency}, \label{eq: reg}
\end{split}
\end{equation}
where {$\alpha$ is a hyperparameter for adjusting the weight of consistency constraint.}

\begin{algorithm}[!t]
    \caption{Unsupervised Training Algorithm.}
    \label{alg:1}
    \begin{algorithmic}
    \STATE \textbf{Input:} original view $V$, view augmenter $g$,  shared information encoder $f$, hyperparameters $\alpha, \beta$, MI estimator $I$
    \STATE \textbf{Output:} the representation of nodes in views $z, z^{'}$
    \end{algorithmic}
    
    \begin{algorithmic}[1]
    \WHILE{$epoch \neq max\_epochs$}
    \STATE $Fix \ g, \ get \  V^{'}=g(V)$
    \STATE $I(f(V);f(V')) + \beta I(V;f(V))$
    \STATE $Update \ f \ to \ minimize \ line \ 3$
    \STATE $Fix \ f$
    \STATE $I(f(V),f(g(V))) + \alpha (\Vert \Lambda- {\Lambda}^{'} \Vert_F^2 + \Vert {C} - {C}^{'} \Vert_F^2)$
    \STATE $Update \  g \ to \ minimize \ line \ 6$
    \ENDWHILE
    \STATE $z, z^{'} = f(V), f(g(V))$
    \STATE \textbf{return} $z, \ z^{'}$
    \end{algorithmic}
 \end{algorithm}

According to Theorem~\ref{Theorem:6}, it is proven that the new pretext task does not corrupt representative information in augmented views during augmentation. 
\begin{theorem}\label{Theorem:6}
    \textit{When applying the InfoMin principle $min \ I(V; V')$ to train the augmenter in LAC, the mutual information between the original view and the augmented view $I(V; V')$ is always greater than zero.}
\end{theorem}

\begin{proof} In the continuous space of LAC, the view $V$ is represented as $(\mathbf{U}, \mathbf{\Lambda}, \mathbf{C})$, and the augmented view as $V'=(\mathbf{U}, \mathbf{\Lambda}', \mathbf{C}')$. The mutual information between the two views is denoted as: 
\begin{equation}
\begin{split}
    I(V; V') &= I((\mathbf{U}, \mathbf{\Lambda}, \mathbf{C}); (\mathbf{U}, \mathbf{\Lambda^{'}}, \mathbf{C'})) \\
    &= I(\mathbf{U};(\mathbf{U}, \mathbf{\Lambda^{'}}, \mathbf{C'})) + I(\mathbf{\Lambda};(\mathbf{U}, \mathbf{\Lambda^{'}}, \mathbf{C'})) \\
    & \quad + I(\mathbf{C}; (\mathbf{U}, \mathbf{\Lambda^{'}}, \mathbf{C'})) \\
    &\geq I(\mathbf{U}; \mathbf{U}) + I(\mathbf{\Lambda}; \mathbf{\Lambda}^{'}) + I(\mathbf{C}; \mathbf{C}^{'})  \\
    &\geq I(\mathbf{U}; \mathbf{U}) > 0.
\end{split}
\end{equation}
The equation demonstrates that the mutual information between the original and the augmented views, $I(V; V')$, is greater than zero. Above all, Theorem~\ref{Theorem:6} is proven.
\end{proof}

\subsection{Pretext Task for the Shared Information Encoder} 
The InfoMax principle is a widely used approach for training encoders. However, training an encoder using only the InfoMax principle in LAC can lead to shortcut problems.
According to Theorem~\ref{theorem:3}, we have:
\begin{equation}
    I(f(V);f(V')) \leq I(V';f(V')),
\end{equation}
\begin{equation}
    I(f(V);f(V')) \leq I(V;f(V)). 
\end{equation}

It shows that the representative information in augmented views utilized by the encoder with the InfoMax is limited by $I(V';f(V'))$ and $I(V;f(V))$. 
Specifically, the encoder $f$ may resort to shortcuts, as it encodes only limited information from augmented views, thereby lowering the upper bound of $I(f(V);f(V'))$. This makes the model easier to train for convergence but ends up with a low-quality encoder.
Therefore, we propose a \textit{sufficiency constraint} that adds a bottleneck loss on top of the mutual information maximization term to exploit the representative information as follows sufficiently:
\begin{equation} \label{eq: bn}
    \mathop{\max}\limits_{f}{\underbrace{I(f(V);f(V'))+\beta(I(V';f(V'))+ I(V;f(V)))}_{sufficiency}},
\end{equation}
where {$\beta$ is a hyperparameter for controlling the weight of bottleneck loss.}

To compute the second term of Eq. \ref{eq: bn}, we follow the setup in \cite{AutoGCL}, and employ an encoder to map the information in $V'$ into an embedding, ensuring that it shares the same dimensions as $f(V')$.

\textbf{Unsupervised Learning.}
The goal of the unsupervised training, guided by the proposed objective InfoBal is to obtain high-quality embeddings. The training process for LAC consists of two stages. 
Initially, we fix the weights of the view augmenter $g$ and use the pretext task based on sufficiency constraint to train the encoder $f$. Subsequently, we train the view augmenter using the pretext task, which integrates both consistency and diversity constraints. These two stages alternate in the training process. The algorithm is illustrated in Alg. \ref{alg:1}.

\begin{table*}[!t]
\caption{The detailed statistics of the dataset.} \label{table:1}
        \begin{center}
            \begin{tabular}{c c c c c c }
            \toprule
                 Category & Dataset & \#Nodes & \#Edges & \#Sparsity & \#Hetero \\
            \hline
                 & Cora  & 2,708 & 5,429 & 0.00074 & 0.19  \\
                 Homo- & CiteSeer & 3,327 & 4,552 & 0.00042 & 0.26 \\
                 geneous & PubMed & 19,717 & 44,324 & 0.00011 & 0.20 \\
                 \& & Photo & 7,650 & 119,081 & 0.00203 & 8 \\
                 Sparse & Computers & 13,752 & 245,861 & 0.00130 & 10 \\
                 & CS & 18,333 & 81,894 & 0.00024 & 15 \\
                 & Phy & 34,493& 247,962& 0.00020 & 5\\
                 \hline
                 Hetero- &Chameleon & 2,277& 36,101& 0.00696 & 0.77\\
                 geneous &Squirrel & 5,201& 217,073& 0.00802 & 0.78\\
                 \hline
                 Dense & Cornell & 183& 295& 0.00880 & 0.89\\
                  & Texas & 183& 309& 0.00922 & 0.89\\
            \bottomrule
            \end{tabular}
        \end{center}
\end{table*}

\section{Experiments}
We begin this section with the introduction of the experimental setup, which includes the details of datasets, evaluation protocols, and baselines. Then we conduct experiments to evaluate LAC by addressing the following research questions:
\begin{itemize}
\item \textbf{$RQ_{1}$. (Effectiveness)} Does LAC perform better than the SoTA GCL frameworks in the unsupervised setting for node classification task?
\item \textbf{$RQ_{2}$. (Generalization Ability)} How well does LAC generalize on various types of graphs and other tasks? 
\item \textbf{$RQ_{3}$. (The necessity of each component)} Are the view augmenter CVA and InfoBal based pretext tasks in LAC both necessary? 
\item \textbf{$RQ_{4}$. (Model Analysis)} Does the effect of MTA and CFA meet the design expectations?
\item \textbf{$RQ_{5}$. (Sensitivity)} Is LAC sensitive to hyperparameters like $\alpha$, $\gamma$, $\tau$, and mask ratio in MTA?
\end{itemize}

\begin{table*}[!t] 
\renewcommand\arraystretch{1.1}
        \caption{The performance of LAC and baselines on seven sparse and homogenous datasets in terms of accuracy in percentage with standard deviations over ten runs. * means the results are quoted from the corresponding papers \cite{GRACE, GCA, MERIT, MVGRL, SFA}. \dag \ means the experiment results are reported based on the open public code. OOM means out of memory. The best performance is highlighted in boldface. The second-best performance is underlined. - means no public code and results can be found.}        
        \label{table:2}    
        \resizebox{\textwidth}{!}{
        \begin{tabular}{c c c c c c c c c} 
            \toprule               
            Category & Model & Cora & CiteSeer & PubMed & CS & Phy & Photo & Computers\\       
            \hline                 
            Graph Represent & RawFeat* \cite{GRACE} & $56.89\pm0.08$&$60.70\pm0.04$&$83.84\pm0.10$&$73.91\pm0.20$&$94.53\pm0.34$&$83.52\pm0.31$& $78.53\pm0.00$\\
              Learning & Node2Vec* \cite{Node2Vec} & $64.50\pm0.51$&$53.05\pm1.02$&$74.29\pm0.39$&$78.03\pm0.39$&$94.66\pm0.15$&$89.61\pm0.24$& $84.39\pm0.08$\\
                & DGI* \cite{DGI} &$83.25\pm0.68$&$ 72.03\pm0.65$&$84.75\pm0.19$&$92.75\pm0.10$&$94.42\pm0.67$&$90.76\pm0.29$& $83.95\pm0.47$\\
                \hline
                Graph Generative & GAE* \cite{VGAE} &$ 76.90\pm0.25$&$60.60\pm0.14$&$82.94\pm0.31$&$90.01\pm0.71$&$94.92\pm0.07$&$91.62\pm0.13$& $85.27\pm0.19$\\
                Learning & VGAE* \cite{VGAE}& $78.92\pm0.46$&$61.22\pm0.11$&$83.07\pm0.29$&$92.11\pm0.09$&$94.52\pm0.01$&$92.20\pm0.11$&$ 86.37\pm0.21$\\
                \hline
                Manual & GRACE* \cite{GRACE} & $83.32\pm0.41$&$72.10\pm0.53$&$85.51\pm0.37$&$91.12\pm0.20$&$95.41\pm0.13$&$92.15\pm0.25$& $87.25\pm0.25$\\
                GCL & GCA* \cite{GCA} &$82.89 \pm0.21$&$72.89\pm0.13$&$85.12\pm0.23$&$93.10\pm0.01$&$95.68\pm0.05$&$92.49\pm0.09$&$87.85\pm0.31$ \\
                Frameworks & MVGRL* \cite{MVGRL} & $83.11\pm0.12$&$\uline{73.33\pm0.03}$&$84.27\pm0.04$&$93.11\pm0.12$&$95.33\pm0.03$&$91.74\pm0.07$& $87.52\pm0.11$\\
                & BGRL* \ \cite{BGRL} &$82.83\pm1.61$&$72.32\pm0.89$&$\uline{86.03\pm0.33}$&$93.31\pm0.13$&$\uline{95.73\pm0.05}$&$93.17\pm0.30$&$ 88.54 \pm 0.03 $\\
                & COSTA \dag \cite{COSTA} &$84.32\pm0.22$&$72.92\pm0.31$&$86.01\pm0.19$&$92.56\pm0.45$&$95.01\pm0.09$&$92.56\pm0.42$&$88.32\pm0.30$ \\
                \hline 
                Statistical  & CCA-SSG* \cite{CCA-SSG} &$84.20\pm 0.40$&$73.10\pm0.30$&$85.39\pm0.51$&$93.31\pm0.22$&$95.38\pm0.06$&$93.14\pm0.14$&$88.74\pm0.28$\\
                GCL Frameworks   & {MC-DCD} \dag \cite{MC-DCD} &-&-&-&$93.60\pm 0.08$&$95.50\pm0.04$&$93.31\pm0.13$&$88.78\pm0.25$ \\
                \hline
                Automated & JOAO-v2 \dag \ \cite{JOAO}&$82.47\pm0.43$&$70.29\pm0.44$&$83.81\pm0.05$&$91.50\pm0.29$&$94.79\pm 0.53$&$91.39\pm0.25$&$86.73\pm0.44$ \\
                GCL & ADGCL \dag \ \cite{ADGCL} &$83.51\pm0.63$&$72.42\pm0.36$&$85.45\pm0.33$&$93.26\pm0.30$&$95.57\pm0.09$&$91.45\pm0.12$& $86.03\pm0.21$\\
                Frameworks & AutoGCL\dag \ \cite{AutoGCL} & $83.86\pm0.25$&$72.62\pm0.50$&$84.27\pm0.52$&$92.37\pm0.24$&$95.15\pm 0.14$&$92.25\pm0.30$&$87.11\pm0.84$ \\
                \hline
                Spectral Framework & GCL-SPAN\dag \ \cite{SPAN} &$\uline{85.51\pm0.61}$ &$72.46\pm0.35$ &  $85.45\pm0.16$  & $\uline{93.76\pm0.38}$ &  OOM &  $92.58\pm0.17$& $\uline{89.96\pm0.13}$ \\
                \hline
                Ours & \textbf{LAC} & $\pmb{86.02}\pm\pmb{0.45}$ & $\pmb{73.51}\pm\pmb{0.62}$& $\pmb{86.05}\pm\pmb{0.21}$ & $\pmb{93.84}\pm\pmb{0.30}$& $\pmb{96.08}\pm\pmb{0.15}$& $\pmb{93.75}\pm\pmb{0.43}$& $\pmb{90.55}\pm\pmb{0.16}$ \\  
            \bottomrule            
        \end{tabular}}
        
\end{table*}

\subsection{Experimental Setup}

\textbf{Datasets.}
We evaluate our approach using seven sparse and homogeneous datasets, including Cora \cite{Planteoid}, CiteSeer \cite{Planteoid}, PubMed \cite{Planteoid}, CS \cite{Coauthor}, Phy \cite{Coauthor}, Photo \cite{Coauthor} and Computers \cite{Coauthor}. We also perform experiments on two heterogeneous graph datasets—Chameleon and Squirrel and two dense graph datasets—Texas and Cornell (all cited from \cite{GeomGCN}). The detailed statistics of the datasets are summarized in Table~\ref{table:1}.


\textbf{Evaluation Protocols.} Following the settings in \cite{ADGCL}, LAC uses a logistic classifier to evaluate the unsupervised trained model for the node classification task. 
We train LAC using Adam Optimizer and apply Xavier initialization uniformly for the modules in the network. We search for the hyperparameter $\alpha$ in different ranges for different datasets, usually between $0.1$ and $1.5$ with a minimal interval of $0.05$. Similarly, for the $\beta$, $\tau$, and mask ratio parameters, we search between $0.1$ and $1.0$ with a minimal interval of $0.05$. The layer of encoder in MTA is in [1,2].
We run the experiments by PyTorch 1.12.0 and Pytorch Geometric 2.1.0. All experiments are conducted on an NVIDIA A6000 GPU (48GB) and an NVIDIA A100 GPU (80GB).

\begin{table*}[!t] 
  \caption{Node classification accuracy in dense graphs, and heterogeneous graphs}
  \label{table:other}
  \centering
  \begin{tabular}{ccccccc}
    \toprule
      &  & manual & statistical& automated& spectral & ours\\
      Dataset & type & BGRL & CCA-SSG & ADGCL & GCL-SPAN & LAC \\
    \midrule
        Squirrel& dense & 39.65$\pm$ 1.88 & \underline{40.07$\pm$1.10} & 39.19$\pm$0.91 & 38.13$\pm$0.17 & \textbf{46.83}$\pm$\textbf{1.38} \\
        Cornell& dense & \underline{56.16$\pm$0.96} & 50.27$\pm$1.65 & 51.57$\pm$6.98 & 51.58$\pm$6.13 & \textbf{72.63}$\pm$\textbf{3.93}\\
    \midrule
        Chameleon& heterogeneous & 55.02$\pm$4.04 & \underline{57.55$\pm$ 2.69} & 55.89$\pm$3.33& 54.50$\pm$1.44 & \textbf{63.66}$\pm$\textbf{0.98}\\
        Texas& heterogeneous & 57.89$\pm$8.80 & 58.90$\pm$1.56 & 63.15$\pm$9.98 & \underline{69.21$\pm$4.31}& \textbf{83.15}$\pm$\textbf{2.10}\\
    \bottomrule
\end{tabular}
\end{table*}

\textbf{Compared baselines.} We mainly compare our LAC\footnote{ \url{https://github.com/linln1/LAC.git}} framework with other SoTA frameworks for node classification. The SoTA methods include six categories: (1) Graph representative learning methods, such as RawFeat \cite{GRACE}, Node2Vec \cite{Node2Vec}, and DGI \cite{DGI}. 
(2) Graph generative learning models such as GAE and VGAE \cite{VGAE}. These models are chosen because LAC contains a generative view augmenter. (3) Manual GCL frameworks include GRACE \cite{GRACE}, GCA \cite{GCA}, MVGRL \cite{MVGRL}, {BGRL \cite{BGRL} and COSTA \cite{COSTA}}. (4) GCL framework based on statistics include CCA-SSG \cite{CCA-SSG} and MC-DCD \cite{MC-DCD}. (5) Automated GCL frameworks contain ADGCL \cite{ADGCL}, JOAO \cite{JOAO}, and AutoGCL \cite{AutoGCL}. (6) Spectral GCL framework contains GCL-SPAN \cite{SPAN}.

\begin{table*}[!t] 
  \caption{Node Clustering task results} \label{table:clustering}
  \scalebox{0.75}{
  \begin{tabular}{ccccccccccc}
    \toprule
     & \multicolumn{2}{c}{BGRL} & \multicolumn{2}{c}{CCA-SSG} & \multicolumn{2}{c}{ADGCL} & \multicolumn{2}{c}{GCL-SPAN} & \multicolumn{2}{c}{LAC} \\
    \midrule
     metrics & NMI & ARI & NMI & ARI & NMI & ARI & NMI & ARI & NMI & ARI \\
    \midrule
        Cora & 0.27$\pm$0.02 & 0.18$\pm$0.02 & 0.50$\pm$0.02 & 0.40$\pm$0.02 & 0.32$\pm$0.03 & 0.20$\pm$0.03 & \underline{0.52$\pm$0.01} & \underline{0.45$\pm$0.01} & \textbf{0.53}$\pm$\textbf{0.01} & \textbf{0.46}$\pm$\textbf{0.01}  \\
        Cornell & 0.16$\pm$0.01 & 0.08$\pm$0.02 & 0.14$\pm$0.01 & 0.06$\pm$0.01 & 0.14$\pm$0.01 & 0.07$\pm$0.02
        & \underline{0.24$\pm$0.01} &  \underline{0.12$\pm$0.01}  & \textbf{0.24}$\pm$\textbf{0.02} & \textbf{0.12}$\pm$\textbf{0.01} \\
        Chameleon & 0.10$\pm$0.01 & 0.05$\pm$0.00 & \underline{0.14$\pm$0.02} & 0.04$\pm$0.01 & 0.11$\pm$0.01 & \underline{0.06$\pm$0.00}  & 0.12$\pm$0.01 & 0.06$\pm$0.00 & \textbf{0.19}$\pm$\textbf{0.01} & \textbf{0.11}$\pm$\textbf{0.01} \\
    \bottomrule
\end{tabular}
}
\end{table*}

\begin{table*}[!t] 
\renewcommand\arraystretch{1.1} 
  \caption{The ablation study results of continuous view augmenter in LAC.}        
        \label{table:3}
  \begin{center}                 
    \resizebox{\linewidth}{!}{
    \begin{tabular}{c c c c c c c c} 
      \toprule               
      Model &Cora&CiteSeer&PubMed&CS&Phy&Photo&Computers\\   
                \hline
                LAC-w/o-MTA & $85.60\pm0.91$ & $72.64\pm0.17$ & $85.76\pm0.14$&$93.15\pm0.27$ & $94.68\pm0.30$& $93.56\pm0.51$&$89.53\pm0.35$ \\
                LAC-EgMsk & $\uline{85.97\pm0.81}$ & $\uline{73.33\pm0.88}$ & $\uline{85.85\pm0.39}$&$93.04\pm0.14$&$95.49\pm0.23$&$\uline{93.62\pm0.59}$&$89.46\pm0.14$\\
                LAC-w/o-CFA & $83.52\pm0.85$ & $69.70\pm1.27$ & $84.59\pm0.10$&$\uline{93.79\pm0.31}$&$95.87\pm0.18$&$93.56\pm0.43$&$\uline{89.95\pm0.44}$\\
                LAC-FtMsk & $85.80\pm0.64$ & $71.73\pm0.95$ &$85.73\pm0.26$&$93.60\pm0.42$&$\uline{95.94\pm0.09}$&$93.25\pm0.26$&$89.44\pm0.19$\\
                \hline
                \textbf{LAC}                
                &$\pmb{86.02}\pm\pmb{0.45}$ & $\pmb{73.51}\pm\pmb{0.62}$& $\pmb{86.05}\pm\pmb{0.21}$ & $\pmb{93.84}\pm\pmb{0.38}$& $\pmb{96.08}\pm\pmb{0.15}$& $\pmb{93.75}\pm\pmb{0.43}$& $\pmb{90.55}\pm\pmb{0.16}$ \\
      \bottomrule            
    \end{tabular}}
  \end{center}
\end{table*}

\subsection{Effectiveness of LAC} 

Table~\ref{table:2} lists the accuracy of node classification using LAC and baselines to learn node embeddings in the unsupervised setting on seven sparse \& homogeneous datasets.
We observe that LAC achieves state-of-the-art (SoTA) accuracy on the node classification task. Specifically, we have the following observations: 

(1) LAC achieves average improvements of 2.45\% and 4.42\% compared to the best graph representation learning method and the best graph generative learning method, respectively. This indicates the effectiveness of LAC as a graph-contrastive learning framework.

(2) The LAC framework demonstrates an average accuracy improvement of 0.76\% over the best manual GCL framework across all seven sparse and homogeneous datasets. It indicates that the LAC framework can replace manual GCL frameworks as it eliminates the need for trial-and-error experiments to find appropriate augmented views and sufficiently extract representative information from those augmented views.

(3) Compared to state-of-the-art automated GCL frameworks, statistical frameworks, and a spectral framework. 
Across the seven sparse and homogeneous datasets, LAC achieves average accuracy improvements of 1.38\%, 0.92\%, and 0.67\%, respectively. For instance, LAC demonstrates an accuracy enhancement of 2.16\% over AutoGCL, 1.82\% over CCA-SSG, and 0.51\% over GCL-SPAN on the Cora dataset. This is because LAC generates appropriate augmented views and uses representative information in graph data sufficiently.

In summary, LAC consistently surpasses state-of-the-art baselines across all datasets. This demonstrates the effectiveness of the LAC framework.



\subsection{Generalization of LAC}
\textbf{The generalization ability of LAC on different datasets.}
Table~\ref{table:other} reports the performance of LAC on heterogeneous and dense datasets with the node classification task, demonstrating its generalization capability to various types of datasets. 
We select Chameleon and Texas as the representatives of heterogeneous datasets, which have heterogeneity scores of 0.77 and 0.89, respectively, and choose Cornell and Texas as dense datasets, whose sparsity is around 0.9\%.
We compare a few of the best models in manual GCL frameworks (BGRL), statistical GCL frameworks (CCA-SSG), automated GCL frameworks (ADGCL), and spectral frameworks (GCL-SPAN) as the baselines. 
The results clearly demonstrate that LAC consistently outperforms baselines across different types of graphs.

\textbf{The generalization ability of LAC on different tasks.} 
We conduct unsupervised node clustering experiments on the Cora, Cornell, and Chameleon datasets to demonstrate LAC's advantages on other tasks. The results are reported in Table \ref{table:clustering}. 
NMI (Normalized Mutual Information score) and ARI (Adjusted Rand Information score) are used to evaluate the performance of node clustering. The results demonstrate that LAC achieves the best performance on different types of graphs. 

\subsection{Ablation Study}

This section presents an ablation analysis of the various components within LAC to demonstrate the necessity of the proposed techniques.

\textbf{The Effectiveness of the Continuous View Augmenter.} To evaluate the effectiveness of the augmentation of both topological and feature information, we have developed two new variants of LAC. These variants are LAC-w/o-MTA, which excludes the MTA module, and LAC-w/o-CFA, which removes the CFA module from the framework. 
To validate the effectiveness of continuous augmentation, we replace MTA in the LAC framework with a random edge masking augmenter, denoted by LAC-EgMsk, and replace CFA with a random feature masking augmenter, denoted by LAC-FtMsk. The results are presented in Table~\ref{table:3}. 


(1) LAC demonstrates superior performance compared to its variants LAC-EgMsk and LAC-FtMsk across all datasets. Specifically, across the seven sparse datasets listed in Table~\ref{table:1}, LAC improves accuracy by an average of 0.43\% and 0.62\% over LAC-EgMsk and LAC-FtMsk, respectively. The results indicate that augmenting topology and feature information in continuous space generates more appropriate augmentation views than discrete augmentation.

(2) LAC performs better than LAC-w/o-MTA and LAC-w/o-CFA in all datasets. On average, LAC is 0.69\% and 1.26\% more accurate than LAC-w/o-MTA and LAC-w/o-CFA, respectively. This demonstrates that both feature augmentation and topology augmentation are necessary.

\begin{table*}[!t] 
\renewcommand\arraystretch{1.1}
	\caption{The ablation study results of training pretext task InfoBal in LAC.}        
	\label{table:4}           
	\begin{center}                 
        \resizebox{\linewidth}{!}{
		\begin{tabular}{c c c c c c c c c} 
			\toprule               
			Tasks &Cora&CiteSeer&PubMed&CS&Phy&Photo&Computers\\     
                \hline
                Max-Max & $\uline{86.02\pm0.65}$ & $72.83\pm0.29$ & $83.27\pm0.28$&$92.83\pm0.45$ & $94.83\pm0.26$& $89.80\pm0.70$ &$82.71\pm3.88$ \\
                Min-Max & $81.10\pm1.94$ & $72.63\pm0.44$ & $83.19\pm0.51$&$ \uline{93.52\pm0.38} $&$90.22\pm1.97$ & $90.35\pm2.19$ & $79.84\pm1.15$\\
                Min-BN & $83.97\pm2.04$ & $73.13\pm0.59$ & $83.54\pm0.39$& $93.52\pm0.32$ & $94.58\pm0.45$ & $92.05\pm0.35$ & $\uline{83.75\pm2.14}$\\
                Reg-Max & $84.85\pm2.43$ & $\uline{73.25\pm0.29}$ &$\uline{84.36\pm0.60}$&$93.04\pm0.36$&$\uline{95.18\pm0.60}$&$\uline{93.62\pm0.05}$ & $82.19\pm3.78$\\
                \hline
                \textbf{InfoBal} & $\pmb{86.02\pm0.45}$ & $\pmb{73.51\pm0.62}$& $\pmb{86.05\pm0.21}$ & $\pmb{93.84\pm0.38}$& $\pmb{96.08\pm0.15}$& $\pmb{93.75\pm0.43}$& $\pmb{90.55\pm0.16}$ \\
			\bottomrule            
		\end{tabular}
    }
	\end{center}
\end{table*}


\textbf{The Effectiveness of the Pretext Tasks based on the InfoBal.}
To validate the effectiveness of the pretext tasks within LAC, we introduce two variants of our pretext tasks. 
To standardize the naming of variants, we adopt the A-B format to describe these variants. Specifically, pretext task A is used to train the CVA, and pretext task B is used to train the encoder.
The first variant, Min-BN, removes regularized loss (consistency constraint) in Eq. \ref{eq: reg} to train the CVA and uses loss in Eq. \ref{eq: bn} to train the encoder. Similarly, the second variant is named Reg-Max, which removes the bottleneck loss in Eq. \ref{eq: reg} for the training of the encoder.
Additionally, we introduce two variants based on commonly used pretext tasks. The Max-Max variant utilizes an InfoMax-based loss to train the CVA and the encoder, respectively. The Min-Max variant employs an InfoMin-based loss to train the CVA while using an InfoMax-based loss to train the encoder.

The results, presented in Table \ref{table:4}, show that the pretext tasks based on InfoBal significantly outperform the other tasks. 
InfoBal's superior performance compared to both Reg-Max and Min-BN illustrates the effectiveness of the consistency constraint and the sufficiency constraint, respectively. In addition, the Max-Max task is better than the Min-Max task on most datasets, which indicates that the Min-Max task excessively destroys consistent information when using InfoMin to generate augmented views, and eventually leads to performance degradation. InfoBal's accuracy has a 4.13\% improvement over the sparse datasets compared to the Min-Max. 
This demonstrates that InfoBal can prevent excessive perturbation of representative information caused by the InfoMin and effectively utilize it to improve the encoder's quality.

To evaluate the effectiveness of the consistency constraint on the overall framework, we compare the performance of the Reg-Max and Min-Max. 
The results show that the Reg-Max improves accuracy by an average of 2.23\% over the Min-Max. 
This suggests that the Min-Max task causes the view augmenter to corrupt excessive representative information during view augmentation, thereby degrading the performance of the GCL framework.
On the other hand, adding the consistency constraint on top of InfoMin allows the augmenter to adaptively control the diversity and consistency during augmentation. It significantly improves the quality of augmented views. 
Therefore, we conclude that the Reg-Max task is more effective than the Min-Max task.

Furthermore, a comparison between the performance of Min-BN and Min-Max reveals that the former achieves an average accuracy improvement of 1.95\% over the latter. It indicates that the sufficiency constraint on InfoMax enables the encoder to extract sufficient representative information from the augmented view.

\subsection{Model Analysis}

\begin{figure}[!t]
    \centering
    \scalebox{0.9}{
        \includegraphics[width=0.9\linewidth]{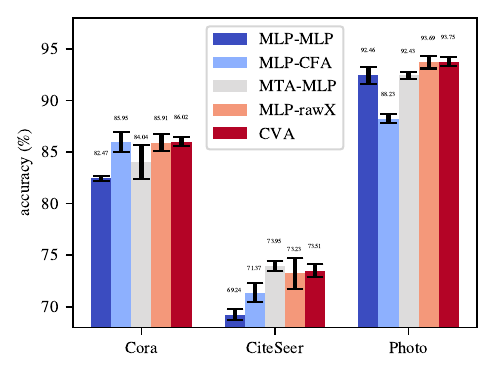}
    }
    \caption{The results of different architectures used in CVA.}
    \label{fig:CVA}
\end{figure}

\textbf{The Analysis of the MTA and CFA designed for CVA.}
We conduct experiments on MTA and CFA to assess whether these carefully designed architectures outperform simpler architectures, such as MLPs. Specifically, we substitute the MTA in the CVA with a two-layer MLP. During the augmentation of views, the MLP does not consider the potential multivariate relationships between different values on the diagonal of the topological information matrix $\mathbf{\Lambda}$ compared to MTA. 
Similarly, we replaced the CFA module with a two-layer MLP, which cannot capture relationships between features in different channels. 
The two variants are named MTA-MLP and MLP-CFA, respectively. 
Furthermore, we replace both modules with two-layer MLP, resulting in the MLP-MLP model. 
To demonstrate the advantages of augmenting feature information within a continuous space, we also directly perturb node features X instead of perturbing C, and we denote this model as MTA-rawX.
Our experimental results are summarized in Figure~\ref{fig:CVA}.

In our study, we compare the accuracy of the CVA with that of MTA-MLP. Across three datasets, CVA exhibits an average accuracy improvement of 0.95\% compared to MTA-MLP. This suggests that the CFA module effectively augments node features by learning and perturbing the distribution of cross-channel node features in continuous space.
Additionally, we compare the performance of the CVA with MLP-CFA. We find that CVA achieved an average accuracy improvement of 2.57\% across the three datasets, indicating the effectiveness of the MTA module design.
Finally, when comparing the original architecture to MLP-MLP, we observe that CVA produces higher-quality augmented views across three datasets, achieving an average accuracy improvement of 3.03\%. 
Additionally, on average, CVA outperformed MTA-rawX by 0.15\%, indicating the advantage of perturbing $\mathbf{C}$ over directly perturbing $\mathbf{X}$. These results validate the effectiveness of the MTA and CFA designs.

\begin{figure}[!tb]
    \begin{minipage}[t]{0.48\linewidth} 
    \centering
    \includegraphics[width=0.9\linewidth]{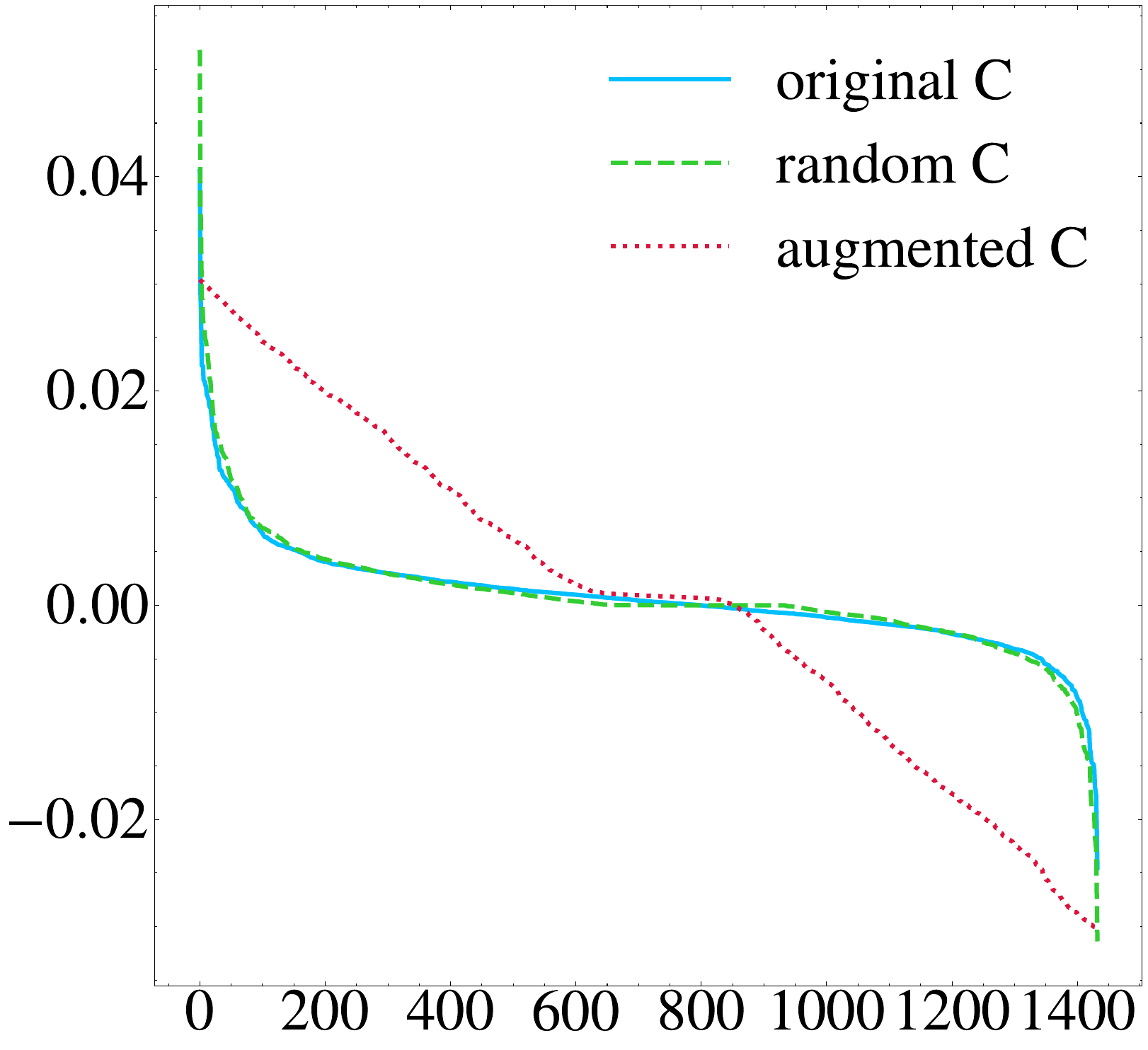}
    \caption{The changes of $\mathbf{C}$ on Cora.} \label{fig:5}
    \end{minipage}%
    \hspace{0.2cm}
    \begin{minipage}[t]{0.48\linewidth} 
    \centering
    \includegraphics[width=0.9\linewidth]{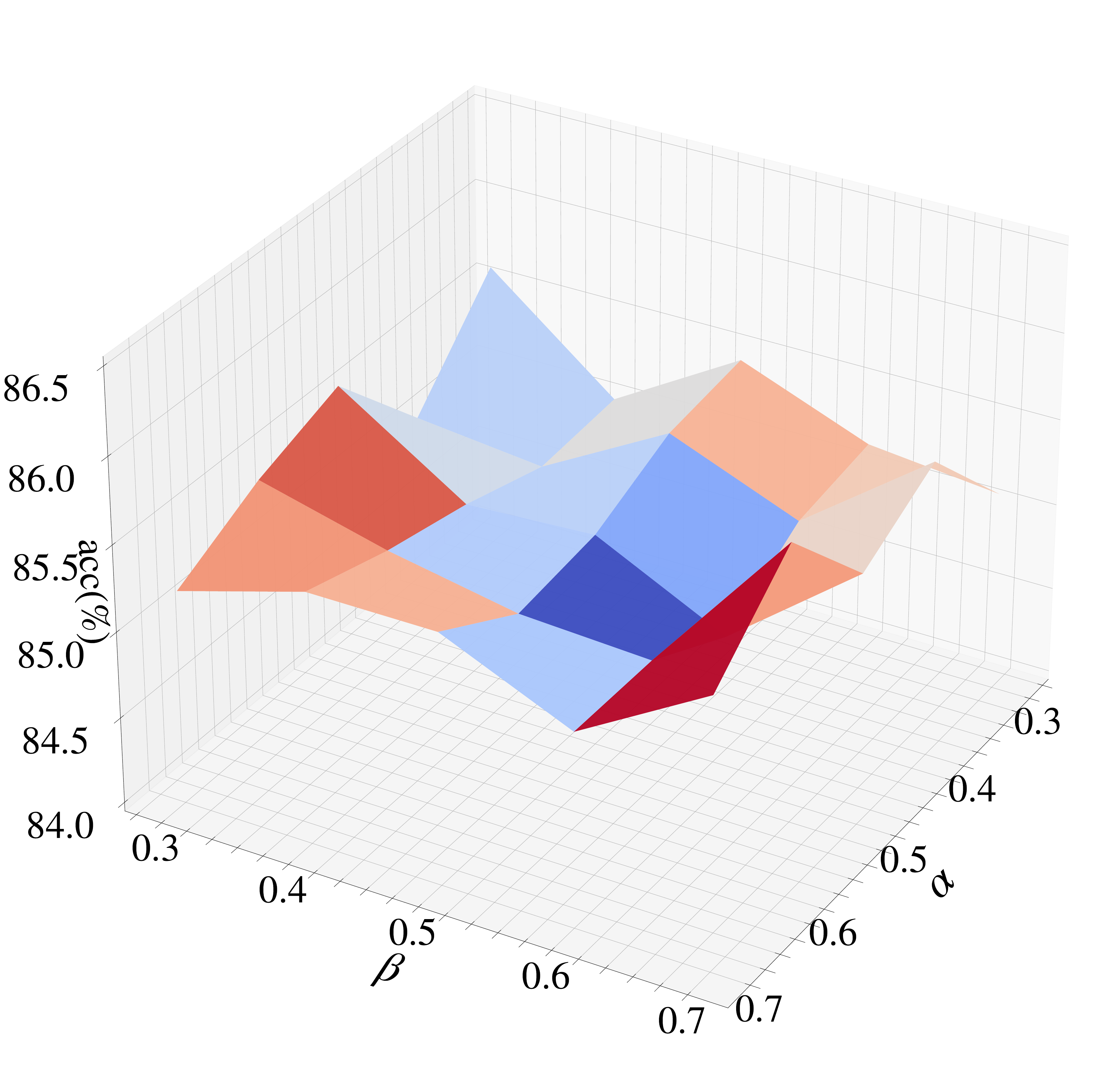}
    \caption{The Impact of different $\alpha$ and $\beta$ on Cora.} \label{fig:6}
    \end{minipage}
\end{figure}

\begin{figure}[!t]
    \centering
    \subfloat[]{\label{a}
    \includegraphics[width=0.23\textwidth]{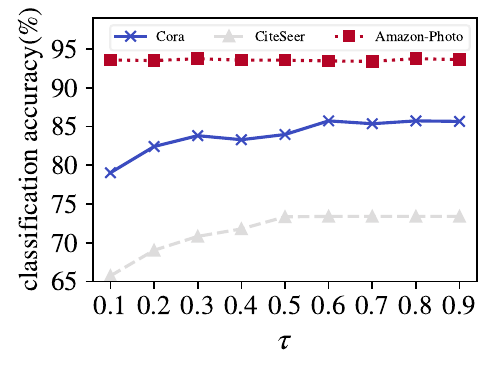}}
    \hfil
    \subfloat[]{\label{b}
    \includegraphics[width=0.23\textwidth]{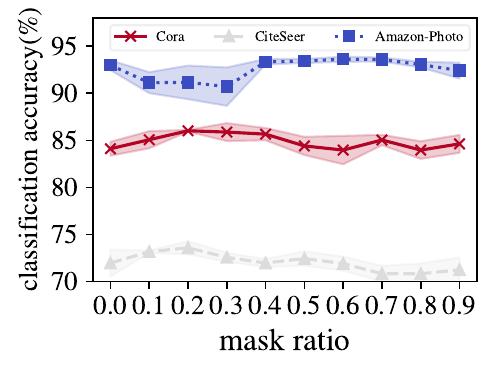}}\\
    \caption{The sensitivity of $\tau$, and mask ratio in LAC.}
    \label{fig:7}
\end{figure}

\textbf{Visual Analysis for augmented node features by CFA.} 
Figure~\ref{fig:5} denotes the value distribution of $\mathbf{C}$ over multiple channels in the Cora dataset.
The horizontal axis represents different channels, namely distinct dimensions of $\mathbf{C}$, while the vertical axis represents the values in corresponding channels. 
In addition to the $\mathbf{C}$ from the original data $\mathbf{X}$ represented by the blue line, we also augment $\mathbf{X}$ using the random feature masking policy with probability equal to 0.2 and calculate $\mathbf{C}$ based on the augmented data, whose distribution is finally represented by the green line. 
Furthermore, we use CFA to automatically learn the augmented $\mathbf{C}$, whose distribution is represented by the red line. The random feature masking results do not improve the unbalanced distribution of original $\mathbf{C}$. On the other hand, the distribution of $\mathbf{C}$ augmented by CFA is much smoother than the original. Thus, CFA can effectively prevent feature collapse caused by uneven distribution \cite{collapse}.

\subsection{Hyperparameters Sensitivity Analysis}

{\textbf{The sensitivity of weights $\alpha$ and $\beta$ in InfoBal.}} Through experiments on Cora, we investigate the impact of the weights of the regularized loss $\alpha$ and bottleneck loss $\beta$ on LAC's performance. The results are reported in Figure~\ref{fig:6}. We observe that the LAC framework exhibits low sensitivity to the two crucial hyperparameters. In other words, when other parameters are fixed, changes in $\alpha$ and $\beta$ do not lead to significant performance variations. Specifically, LAC's performance remains stable on the Cora dataset when $\alpha$ and $\beta$, with other parameters held constant. The highest accuracy performance is only 1.15\% higher than the lowest.

\textbf{The $\tau$ in InfoNCE.} In Figure~\ref{fig:7}\subref{a}, we depict the model accuracy with variations in the temperature parameter $\tau$ in InfoNCE. The model's accuracy ranges from 79.04\% to 85.73\% on Cora and from 93.41\% to 93.75\% on Amazon-Photo as the temperature changes. A marginal accuracy change of 1.52\% is observed on CiteSeer. Based on these observations, We conclude that the performance of LAC is not significantly affected by modifications to $\tau$ as $\tau$ increases beyond 0.5.

\textbf{The mask mechanism in MTA.} In Figure~\ref{fig:7}\subref{b}, the impact of the hyperparameter mask ratio on the MTA is illustrated. Mean-variance curves are utilized to visualize the results, with shaded areas indicating variance. On the Cora dataset, setting the mask ratio to 0 results in a 0.96\% accuracy loss compared to the accuracy at $\tau=0.1$, accompanied by a significant variance. This suggests that the mask mechanism significantly influences LAC's accuracy in specific datasets. However, when increasing the mask ratio from 0.1 to 0.9, the framework's average accuracy variation remains below 2.0\%. This indicates that LAC is not sensitive to changes in the mask ratio.

\section{Related Works}

\subsection{Data Augmentation Methods in GCL}
Graph data augmentation encompasses two primary aspects: topology and features. For topology, traditional methods involve manually selecting probabilities for node deletion \cite{AutoGCL}, edge perturbation \cite{GRACE}, subgraph selection \cite{GraphCL}, and graph diffusion \cite{MVGRL} based on trial-and-error experiments. Recent works have shifted towards automated GCL frameworks that optimize augmentation strategies strategies through data-driven techniques. JOAO \cite{JOAO} utilizes min-max optimization to determine the optimal weights for various discrete data augmentation methods.
AutoGCL \cite{AutoGCL} learns the probability of node masking from the data. ADGCL \cite{ADGCL} learns the Bernoulli distribution probability for each edge. GPA \cite{GPA} learns the optimal combination weights of discrete data augmentation methods for each graph based on JOAO. Ada-MIP \cite{AdaMIP} uses hybrid augmentation strategies. However, the above methods all use a discrete way to augment the topology information.
For feature information, the existing work usually masks certain node features or dimensions  \cite{GRACE, MVGRL, JOAO}. These methods do not continuously change the value of the node feature. The random projection of the feature matrix by COSTA is equivalent to adding an orthogonal continuous unbiased noise to the feature matrix. Through SVD decomposition of the feature matrix and rebalancing of the singular values, SFA \cite{SFA} generates a continuous perturbation of the feature information, thereby avoiding dimension collapse. However, they still overlook the continuous augmentation of topological information.

\subsection{Pretext Tasks in GCL}
{The pretext tasks within the GCL framework are categorized into two main types \cite{MC-DCD}. The first type is based on information theory principles. In this category, mutual information (MI) estimators calculate MI across multiple views or representations. The most commonly used principle is InfoMax \cite{InfoMax}, which aims to maximize consistency in the representation across various views. Conversely, InfoMin \cite{InfoMin} seeks to enhance the diversity of information between views in unsupervised settings \cite{ADGCL, GraphCLA, AutoGCL}. 
They ignore the consistency of information between multiple views and finally produce completely different augmented views, which is a shortcut solution. 
The second category of pretext tasks are designed based on statistics theory, such as CCA-SSG \cite{CCA-SSG} and MC-DCD \cite{MC-DCD}. These GCL frameworks use statistical metrics instead of MI estimators. CCA-SSG applies canonical correlation analysis \cite{CCA} to graph data, while MC-DCD identifies optimal statistical indicators from many candidates for different datasets.}

\section{Conclusion}
GCL is an effective methodology for learning latent information in graph data. However, existing GCL frameworks generate inappropriate augmentations and utilize the representative information in graph data insufficiently. 
We proposed a novel framework for GCL called LAC. The CVA within LAC simultaneously augments topology and feature information in an orthogonal continuous space, thereby generate appropriate augmented views. 
The InfoBal pretext task applied in LAC sufficiently utilizes the representative information in the graph data by adding a consistency constraint on InfoMin and a sufficiency constraint on InfoMax.
Our experimental results demonstrate that LAC significantly outperforms existing GCL frameworks.

\bibliographystyle{unsrt}  
\bibliography{references}  

\end{document}